\RequirePackage{etoolbox}
\csdef{input@path}%
{%
 {sty/},
 {img/},
}
\documentclass{elsevierbook}

\let\AfterPackage\undefined
\usepackage[square,sort,comma,numbers]{natbib}
\usepackage{times}
\usepackage{epsfig}
\usepackage{graphicx}
\usepackage{amsmath}
\usepackage{amssymb}
\usepackage{color}
\usepackage{parskip}
\usepackage{url}
\usepackage{nomencl}
\usepackage{multirow}
\usepackage{multicol}
\usepackage{hhline}
\usepackage{color}
\usepackage{calc}
\usepackage{booktabs}

\usepackage{tabularx}

\begin{document}


\Mainmatter
  \begin{frontmatter}
\newcommand{\shan}[1]{\textcolor{blue}{#1}}
\chapter{Robotic Perception of Object Properties using Tactile Sensing}\label{chap1}
\begin{aug}
\author[addressrefs={ad1}]%
{%
\fnm{Jiaqi} \snm{Jiang}%
}%
\author[addressrefs={ad1}]%
{%
\fnm{Shan} \snm{Luo}%
}%
\address[id=ad1]%
{%
smARTLab,
Department of Computer Science,
University of Liverpool,
United Kingdom. \\
Emails: \{jiaqi.jiang, shan.luo\}@liverpool.ac.uk.
}%

\end{aug}
\begin{abstract}
The sense of touch plays a key role in enabling humans to understand and interact with surrounding environments. For robots, tactile sensing is also irreplaceable. While interacting with objects, tactile sensing provides useful information for the robot to understand the object, such as distributed pressure, temperature, vibrations and texture. 
During robot grasping, vision is often occluded by its end-effectors, whereas tactile sensing can measure areas that are not accessible by vision. 
In the past decades, a number of tactile sensors have been developed for robots and used for different robotic tasks. 
In this chapter, we focus on the use of tactile sensing for robotic grasping and investigate the recent trends in tactile perception of object properties. We first discuss works on tactile perception of three important object properties in grasping, i.e., shape, pose and material properties. We then review the recent development in grasping stability prediction with tactile sensing. Among these works, we identify the requirement for coordinating vision and tactile sensing in the robotic grasping. To demonstrate the use of tactile sensing to improve the visual perception, our recent development of vision-guided tactile perception for crack reconstruction is presented. In the proposed framework, the large receptive field of camera vision is first leveraged to achieve a quick search of candidate regions containing cracks, a high-resolution optical tactile sensor is then used to examine these candidate regions and reconstruct a refined crack shape. 
The experiments show that our proposed method can achieve a significant reduction of mean distance error from 0.82 mm to 0.24 mm for crack reconstruction. Finally, we conclude this chapter with a discussion of open issues and future directions for applying tactile sensing in robotic tasks.

\end{abstract}

\begin{keywords}
\kwd{Tactile sensing, robotic perception, robot grasping, material recognition, shape recognition, pose estimation, grasping stability analysis.}
\end{keywords}

\end{frontmatter}

\section{Introduction}\label{sec1}
Humans explore the environment in their close vicinity using rich sensory information such as the visual information obtained from their eyes and the tactile feeling through physical interaction with cutaneous receptors. Among these sensing modalities, the sense of touch is a key source of information for predicting object properties. From touch sensing, diverse sensory information can be obtained, such as pressure, vibration, pain and temperature.

For robots, tactile sensing is also an irreplaceable modality. Compared to vision and other sensing modalities, tactile sensing is superior at processing material characteristics and detailed shapes of objects \cite{dahiya2012robotic,lederman2009haptic,luo2017robotic}.
By analyzing the data from tactile sensors, the properties of the object in contact can be extracted such as distributed pressure, vibrations and surface textures. Much of such information is inaccessible for remote sensors like cameras due to the occlusion of the end-effectors, which makes tactile sensing highly important in grasping and manipulation. Furthermore, tactile sensing is not subject to factors like light conditions that the performance of vision heavily depends on.

To enable robots to have the sense of touch as humans, in the past decades researchers in the field of robotics have put many endeavours in the development of tactile sensing solutions~\cite{lepora2019pixels,luo2017knock,luo2018vitac,gomes2020geltip,gomes2020blocks,cao2021touchroller,xie2013fiber,luo2021vitac} for the perception of object properties. In this chapter, we review the recent development in robotic perception of object properties with tactile sensing. Three types of object properties that are important for robot grasping tasks are investigated: the material, shape and pose of the object. The grasping stability prediction using tactile sensing is also discussed, which is vital in object grasping.

 Through the survey, we find that many robotic tasks, especially dexterous grasping and manipulation, can only be performed with vision and tactile sensing together. To demonstrate the coordination of the visual and tactile perception, our recent work in vision-guided tactile crack perception~\cite{jiang2021vision} is introduced. In this work, camera vision is first used to quickly search the candidate crack regions, a high-resolution optical tactile sensor is then applied against these candidate regions. Finally, a refined crack shape is reconstructed from the obtained tactile images. In this task, vision shows a large field of view and is able to detect the potential crack areas. On the other hand, tactile sensing can enhance the fineness of the reconstructed crack shape thanks to the characteristics of being less susceptible to light and noise. To evaluate the proposed method, we collected a dataset by investigating 10 mock-up structures that were manufactured using 3D printing. The extensive experiment results demonstrate that compared to only using vision itself, our proposed approach achieves a significant reduction of mean distance error from 0.82 mm to 0.24 mm for crack reconstruction. Furthermore, the proposed method is more than 10 times faster than passive tactile perception in terms of tactile data collection time.
%

%



The remainder of this chapter is organized as follows: the works on material properties recognition using tactile sensing are first discussed in Section~\ref{1.2}, followed by discussions of object shape estimation and object pose estimation via touch, and grasping stability prediction in Sections~\ref{1.3}, Section~\ref{1.4} and Section~\ref{1.5}, respectively. The vision-guided tactile perception for crack reconstruction is then introduced in Section~\ref{1.6}. At the end of this chapter, a conclusion is given along with the discussion on open issues and potential future directions. 



\section{Material properties recognition using tactile sensing} \label{1.2}
To effectively grasp and manipulate objects, robots need to recognize the materials of the objects that they are interacting with. The material properties of an object include roughness, softness and hardness etc. By recognizing the object's material properties, the robot can infer a reasonable grasping force and further achieve effective interaction with an object, e.g., grasping a fragile glass cup without damaging it. Furthermore, the understanding of material properties enables robots to facilitate the task. For example, in a task of laundry sorting, by recognizing the materials of different garments with tactile sensing (and also vision), the robot will be able to separate the items, select appropriate laundry options and prevent potential damage to the clothes.
In general, there are two types of approaches for material recognition using tactile sensing: texture based and stiffness based. 

\textbf{Texture based tactile material recognition.}
Surface texture information is usually closely related to the friction coefficients, roughness, and micro-structure patterns of objects.
Among early studies, hand-crafted features such as force variation~\cite{sinapov2011vibrotactile,liu2012surface} were widely used. In~\cite{sinapov2011vibrotactile}, the vibrations detected by the accelerometer were first transferred to the frequency-domain by Fast Fourier Transform (FFT), then two machine learning algorithms, i.e., Support Vector Machine (SVM) and k-Nearest Neighbors (k-NN) were used to infer the surface texture class. In~\cite{liu2012surface}, the surface physical properties were determined by the application of a dynamic friction model while a finger-shaped force/torque sensor sliding along the object surface at varying speeds. In~\cite{oddo2011roughness}, a microelectromechanical system (MEMS) based tactile sensor was developed and used to discriminate the roughness of object surfaces based on the estimated vibratory patterns during the sliding motion. In~\cite{kim2005texture}, a MEMS-based tactile array sensor was employed to distinguish simple textures using Maximum Likelihood (ML) estimation. In~\cite{liu2016object}, a joint kernel sparse coding method was developed, and this method can explicitly take the intrinsic relations between tactile sensors of BarretHand into account. 

Instead of using electronic tactile sensors that were used in the above works, some studies classified surface textures with optical tactile sensors~\cite{lee2019touching,luo2018vitac}. In such sensors, cameras are used to capture the deformation of a silicone layer that is placed on the top of the camera. Thanks to the use of a camera in the optical tactile sensors, detailed textures of the object can be captured. In~\cite{li2013sensing}, the camera-based GelSight sensor~\cite{johnson2011microgeometry} was used to obtain the tactile images by pressing the sensor against different objects. An algorithm named Multi-scale Local Binary Pattern (MLBP) was proposed to classify different surface textures. Similarly, another camera-based tactile sensor TacTip~\cite{5174720} was used to analyze the object textures in~\cite{winstone2013tactip}. 

Neural networks have recently been used as feature learners to infer material properties. In~\cite{baishya2016robust}, the raw 24,000-dimensional sensor signal was fed to a convolutional neural network and a high classification accuracy of up to 97.3\% was achieved. In~\cite{taunyazov2019towards}, convolutional and recurrent neural networks were applied to learn feature representations from raw data acquired with the hybrid touch and sliding movements. On the other hand, the use of deep neural networks may result in high computational costs and make it challenging to deploy the learned models in real-time operations.

To enhance the efficiency of the texture classification, a spiking neural network was used to can achieve a high classification accuracy comparable to Artificial Neural Network (ANN) with a faster inference in~\cite{taunyazovfast}.
In~\cite{9341333}, a spatio-temporal attention model was proposed for the material recognition by using the tactile texture sequence, where the spatial attention was used to pay more attention to the dominant features for each tactile frame, and temporal attention was used to model the relevance of different regions in the whole contact event. As a result, redundant features can be suppressed, and the salient features can be extracted effectively for the material recognition. 



\textbf{Stiffness based tactile material recognition.}
Object stiffness is another cue that can be used for material property recognition~\cite{nanayakkara2016stable}. 
By using a BioTac~\cite{wettels2008biomimetic} sensor, the object compliance, i.e., the reciprocal of stiffness, can be estimated either using the contact angle of the fingertip~\cite{xu2013tactile} or investigating the BioTac electrode data~\cite{su2012use}. In~\cite{drimus2014design}, a piezoresistive tactile sensor was used, and the objects were classified by applying a KNN classifier as well as the dynamic time warping algorithm~\cite{sakoe1978dynamic} that can calculate the distance between time series.
The camera-based GelSight sensor was also applied to object stiffness estimation~\cite{yuan2016estimating}. 
In~\cite{yuan2016estimating}, the object hardness was estimated based on a numerical model that compared the contact area changes and contact surface geometry as well as the normal force value during pressing the GelSight sensor.

The object materials can be recognized by using different tactile sensors and different sensing cues. As summarized in Table~\ref{tab:re1}, there are different motions used in texture based and stiffness based material recognition methods: data of acceleration, force vibration and acoustic data can be obtained via the motions of sliding, tapping and scratching to reveal the surface textures, whereas data of force variances can be acquired via the motions of squeezing, knocking and pressing to reveal the stiffness of the object. When an optical tactile sensor is used, cues of both object surface textures and stiffness can be obtained by analyzing the tactile images. The motions employed and resultant data types also result in pros and cons of each method: the use of cameras in the optical tactile sensors can capture the detailed micro-structure patterns of object textures in one high-resolution tactile image, whereas it also comes along with high computational cost compared to the use of other tactile sensors.


\begin{table}
\centering
\caption{A summary of material recognition methods with touch sensing}
\def\arraystretch{1.5}
\begin{tabular}{p{1.8cm} | p{2.3cm} | p{2.3cm}| p{3cm} } 

\textbf{Method Type}& \multicolumn{2}{c|}{Texture based} & Stiffness based \\ 
\hline
\textbf{Motions} & Sliding, tapping, scratching & Pressing & Squeezing, knocking, pressing \\
\hline 
\textbf{Data types} & Acceleration, force vibration, acoustic data & Tactile images & Force variances, tactile images \\
\hline 
\textbf{Advantages} & Low cost and limited computational expenses & Micro-structure patterns of object textures can be captured in one tactile image & Limited computational costs; complementary to other information \\

\hline
\textbf{Disadvantages} & Interaction actions like scratching may damage objects & Low resolution of tactile sensors may cause difficulties for processing & Interaction actions like knocking may damage objects \\

\hline
\textbf{References} & \cite{sinapov2011vibrotactile},  \cite{liu2012surface} & \cite{li2013sensing},  \cite{winstone2013tactip} &  \cite{drimus2014design}, \cite{kaboli2014humanoids}\\

\end{tabular}
\label{tab:re1}
\end{table}

\section{Object shape estimation using tactile sensing} \label{1.3}
The ability to identify or reconstruct the shape of objects is crucial for robots to perform multiple tasks such as grasping and in-hand manipulation. By gathering the shape information of objects, robots can better plan and execute grasping strategies and trajectories. There are two main approaches to achieve shape estimation, i.e., shape recognition and shape exploration, and we will discuss them below.

\textbf{Tactile shape recognition.}
Due to the limitation of the low resolution, in most early works tactile sensors were used to collect contact points, and generate point clouds to represent object shapes~\cite{allen1988haptic}. The mosaics of tactile measurements~\cite{pezzementi2011object} were used to reconstruct object global geometry. Such approaches are time-consuming, especially when a large object surface needs to be explored. 

Instead of collecting contact points or local measurements, the object shapes can be represented by extracting features from tactile readings~\cite{luo2015tactile}. In~\cite{luo2015novel}, a new tactile-SIFT descriptor was proposed to extract features in the tactile image and 91.33\% classification accuracy on shape recognition was achieved. In~\cite{luo2016iterative, luo2019iclap}, a novel method named Iterative Closest Labeled Point (iCLAP) was introduced that fundamentally linked kinesthetic cues (contact points) and tactile features extracted from local patterns, and showed superior object recognition performance compared to methods using either contact points or the local tactile features only. 
When a tactile sensor interacts with an object, two types of cues are obtained: local features of the object that indicate the local shapes of the object and the distribution of the contact points that gives a global view of the object in the 3D space.

In the above methods, either local contacts or local features are gathered to train a model firstly, then the trained model is used to recognize the objects in the test set by matching the new observations with the ones in the training. To this end, a database of the object shapes needs to be built. Such methods show a good performance in recognizing object shapes that are known in the database. However, when the object shape is not available in the database, tactile shape exploration is needed.



 

\textbf{Tactile shape exploration.}
To reduce the dependence on prior knowledge such as object shapes, researchers have proposed several approaches that reconstruct object shape through shape exploration~\cite{luo2015localizing}, particularly in the paradigm of active shape exploration. In~\cite{lepora2017exploratory}, a probabilistic approach based on Bayes’ rule was proposed to explore the object's surface through rotating the tactile sensor and moving the sensor tangentially over the object surface. To improve the generalization performance of shape exploration, a deep convolutional neural network was designed in~\cite{lepora2019pixels} to replace the probabilistic model in~\cite{lepora2017exploratory}. Nevertheless, the performance of those methods heavily depends on the localization error of end-effector and requires a long exploration time.
Hence, some other studies use Gaussian Process Implicit Surface (GPIS)~\cite{williams2007gaussian} as object shape representations~\cite{el2013generation,jamali2016active,driess2017active,driess2019active} to reduce the impact of localization uncertainty and accelerate the exploration. 

In~\cite{el2013generation,jamali2016active}, active touch probing was performed at discrete query points. To avoid inefficient touch-and-retract motions, an active learning framework was proposed in~\cite{driess2017active} based on optimal query paths to address the efficiency issue of tactile shape exploration. Furthermore, Driess et al.~\cite{driess2019active} extended the previous work~\cite{driess2017active} and used multiple end-effectors at the same time to achieve a more efficient exploration.

However, the objects used in the above works were fixed on the table to avoid the movement of objects, which makes it hard to be applied in real grasping tasks. To address the uncertainties of movable objects, some studies attempted to recover a movable object's shape by using a series of contacts~\cite{yu2015shape,suresh2020tactile}. Inspired by the paradigm of Simultaneous Localization And Mapping (SLAM) in mobile robotics, an approach that uses contact measurements and planar pushing mechanics as constraints in batch optimization was proposed in~\cite{yu2015shape}. Due to the usage of ordered control points as object shape representations, the method proposed in~\cite{yu2015shape} is not suitable for exploring objects of arbitrary shapes, and will fail easily when an incorrect data-association exists. To address this issue, a recent work~\cite{suresh2020tactile} expanded the method in~\cite{yu2015shape} by combining efficient GPIS regression with factor graph optimization over geometric and physics-based constraints. The results showed a promising localization and shape estimation performance in both simulated and real-world settings. The localization and exploration of the contact on the object can also be facilitated by matching the tactile features with a visual map of the object. In~\cite{luo2015localizing}, the visual-tactile localization problem is also treated as a probabilistic estimation problem for the first time, i.e., a SLAM problem, that was solved in a framework of recursive Bayesian filtering.


As summarized in Table~\ref{tab:re3}, most works in tactile shape recognition focus on the shape estimation of fixed objects, which simplifies the task by ignoring the object's dynamics models. Only a few works can estimate the shape of a movable object, whereas they could only be applied in two-dimensional shape estimation. As for the shape representations, there are three different ones: shape class, ordered points and GPIS. The different object states and shape representations also result in pros and cons of each method: those methods using shape classes as the representations are only applicable to specific scenarios such as assembly automation, whereas they are able to perform the perception much faster than those methods using points and GPIS as shape representations. Tactile object shape exploration methods are less dependent on priors and therefore are more suitable for the estimation of unseen objects, ones of arbitrary shapes and even movable objects.


\begin{table}
\centering
\caption{A summary of the methods for tactile object shape estimation}
\def\arraystretch{1.5}
\begin{tabular}{p{1.8cm} | p{1.8cm} | p{1.8cm} | p{1.8cm}|p{1.8cm}} 

\textbf{Method Type}& Shape recognition & \multicolumn{3}{c}{Shape exploration}  \\ 
\hline
\textbf{Object State} & Fixed & Fixed & Fixed & Movable \\
\hline 
\textbf{Shape Representation} & Shape class & Ordered points & GPIS & Ordered points or GPIS \\
\hline 
\textbf{Advantages} & Low computational cost & High accuracy & Suitable for 3D exploration &  Suitable for movable objects\\
\hline
\textbf{Disadvantages} & Need some prior shape information, not suitable for unseen objects & Long exploration time and performance will drop when explorating 3D objects & The computational effort is not suitable for real-time tasks & Long exploration time and can only reconstruct 2D shapes \\

\hline
\textbf{References} & \cite{pezzementi2011object}, \cite{luo2015novel}, \cite{luo2016iterative} & \cite{lepora2017exploratory}, \cite{lepora2019pixels} & \cite{driess2017active}, \cite{driess2019active} & \cite{yu2015shape}, \cite{suresh2020tactile}\\

\end{tabular}
\label{tab:re3}
\end{table}

\section{Object pose estimation using tactile sensing} \label{1.4}
For dexterous manipulation, it is essential to estimate the pose of the object in hand accurately. A small estimation error of object pose could result in misplacing robotic fingers on the object and even damaging the object. Recently, pose estimation is an important topic in the field of computer vision. However, during object manipulation, the gripper or the arm of the robot may partially or completely occlude the object from vision, which makes the visual pose estimation not stable anymore. To address this problem, tactile sensing can be used as a complementary modality for determining the pose of an object in grasping and manipulation. 

Particle filtering has been extensively used in vision-based robot localization problems. In the past years, it has also found its applications in tactile object pose estimation~\cite{corcoran2010measurement,platt2011using,vezzani2017memory,von2020contact}. In~\cite{corcoran2010measurement}, particle filters were applied to estimate a tube's pose, in which some parts of the robot's fingers were not in contact with the tube. In~\cite{platt2011using}, a particle filtering approach using proprioceptive and tactile measurements was proposed to localize the small objects, e.g., button or grommet, embedded in a flexible material such as thin plastic. In~\cite{vezzani2017memory} an approach named Memory Unscented Particle Filter (MUPF) was proposed that localizes objects recursively in real-time. In~\cite{von2020contact}, a particle-filter based method was proposed to estimate in-hand objects pose by bringing the object into contact with a surface.

In some other works, the pose of the in-hand object was estimated by matching geometry with a pressure sensor~\cite{bimbo2016hand} or a camera-based tactile sensor~\cite{li2014localization,bauza2019tactile}. In~\cite{bimbo2016hand}, the in-hand object pose was estimated by matching Principal Component Analysis (PCA)-based features extracted from tactile pressure readings to the object’s geometric features. In~\cite{li2014localization}, a feature-based height map registration method was proposed to localize the small objects, e.g., a USB cable.
Bauza et al. in~\cite{bauza2019tactile} used CNNs to compute heightmaps and a coarse-to-fine strategy for localizing objects.
With this approach, each object has to be explored individually to determine the tactile map of its global shape.

In summary, particle filtering and feature matching have been used to achieve tactile object pose estimation. However, the pose estimation problems are still not fully solved yet. For example, most particle filter based approaches are sensitive to parameters of models and need a great number of iterations to infer the estimated results. In contrast, feature-based matching methods for pose estimation have to obtain a tactile map for each object and extensively explore the object, which is time-consuming.

\section{Grasping stability prediction using tactile sensing} \label{1.5}
Apart from the material, shape and pose of the object, it is also important to predict the grasping stability with tactile sensing while they are being grasped or manipulated. With the prediction of grasping stability, a robot can find a stable grasping pose and prevent objects from falling. The prediction methods can be summarised into two types: slip detection and grasping stability regression, and we will discuss them as follows.

\textbf{Tactile slip detection.}
Some researchers simplify the grasping stability prediction as a slip detection task, when the object is being lifted. In~\cite{melchiorri2000slip}, a force/torque sensor and a matrix-based tactile sensor were used to detect both translational and rotational slip with the prior of the contact surface's frictional coefficient. In~\cite{kondo2011development}, a fabric sensor of woven electroconductive yarns was used to detect the slip by detecting the change of resistance that depends on the yarn stretch. Although the above method showed good performance in grasp stability prediction, the need for prior knowledge like the frictional coefficient in~\cite{melchiorri2000slip} and the initial stretch of yarn in~\cite{kondo2011development} pose a restriction on their applications.


To predict the slip between the object and the hand without the priors, 
a 3-layer neural network was used in~\cite{su2015force} to predict the slip, which achieves a classification accuracy of 80\%. 
However, as the data in~\cite{su2015force} was squeezed into a one-dimensional array, the spatial distribution inherent in the tactile sensor was lost. To exploit the spatial relationships between different electrodes on the BioTac sensor, in~\cite{garcia2019tactilegcn} the Graph Neural Networks (GNNs) were used which yielded a validation accuracy of 92.7\% on predicting the grasp stability of novel objects.

Tactile images obtained from optical-based tactile sensors have also been used to detect the slip in recent years. In~\cite{dong2017improved}, a two-fingered gripper composed of two GelSight sensors was used to detect slip events by measuring the relative displacement between object texture and markers. In~\cite{james2018slip,james2020slip}, two other different optical tactile sensors, i.e., TacTip \cite{ward2018tactip} and T-MO \cite{james2020tactile}, were employed for detecting slip. The results in~\cite{james2018slip,james2020slip} showed that the slip is detected effectively employing an SVM classifier with the movement of pins as features.



\textbf{Tactile grasping stability regression.}
Different from slip detection that is used to prevent in-hand objects from dropping, grasping stability regression is normally used for picking up objects with a robot arm. To assess the grasp stability, multiple works \cite{bekiroglu2011learning,bekiroglu2013probabilistic,krug2016analytic} used the image moments of tactile readings as features in their classifiers. In~\cite{bekiroglu2011learning,bekiroglu2013probabilistic}, a kernel-logistic regression model was presented. In the model, pose and touch are the conditional factors for grasping probability.   
To reduce the modeling effort and computational load, a wrench-based reasoning method was proposed in~\cite{krug2016analytic} which achieved comparable performance with \cite{bekiroglu2011learning,bekiroglu2013probabilistic}.
In~\cite{li2018slip}, CNNs were combined with Long Short-Term Memory Networks (LSTMs) for predicting the grasp stability.

Though promising results have been achieved, the above works can only estimate the stability of an ongoing grasp with tactile readings and are not capable of regrasping the object when a failure of grasp happens. The tactile readings can also be used for computing the action level information~\cite{li2020review}. Inspired by the actions that humans can adjust hand posture to perform dexterous grasping by using only tactile information, some researchers have investigated assessing current grasp and selecting grasp adjustments to produce a stable new grasp simultaneously.  
In~\cite{chebotar2016self}, the grasp outcome estimated with a grasp stability predictor was used as a reward signal that supervises and provides feedback to the regrasping reinforcement learning algorithm. In~\cite{calandra2018more}, an action-conditional model using tactile readings of GelSight sensor as input was proposed to predict the outcome of a candidate grasp adjustment. In contrast to~\cite{chebotar2016self}, the action-conditional model presented in~\cite{calandra2018more} learned the actions entirely from raw inputs. Similar to~\cite{calandra2018more}, a re-grasping approach was presented in~\cite{hogan2018tactile} using a camera-based tactile sensor. It used CNNs to predict grasp quality, and made local adjustments through simulating rigid-body transformations of tactile readings.

As summarized in Table~\ref{tab:re2}, slip detection and grasping stability regression are used for different application scenarios. Slip detection methods are normally used for preventing in-hand or lifted objects from dropping, whereas grasping stability regression methods are usually used to select the best grasp proposal for picking up the ungrasped objects. 
Moreover, tactile grasping stability regression has a wider application prospect compared to tactile slip detection. For example, the grasp outcome from tactile grasping stability regression can be used to adjust grasp proposals and secure the object grasp. 



\begin{table}
\centering
\caption{A summary of methods for tactile grasping stability prediction}
\def\arraystretch{1.5}
\begin{tabular}{p{2cm} | p{3.5cm} | p{3.5cm}} 

\textbf{Method Type}& Slip detection  & Grasping stability  regression \\ 
\hline

\textbf{Object States} & In-hand or lifted & Ungrasped  \\
\hline 


\textbf{Application scenarios} & To prevent in-hand or lifted objects from dropping & To pick up objects and prevent them from dropping  \\
\hline 

\textbf{Advantages} & Low-cost computation, generalized to novel objects & Can be used for ungrasped objects, and adjust hand posture to produce a stable new grasp based on ongoing grasp \\

\hline
\textbf{Disadvantages} & Not suitable for ungrasped objects, and may need the prior knowledge of frictional coefficients & Low resolution of tactile sensors may cause difficulties for processing, and may need the prior knowledge of object models \\

\hline
\textbf{References} & \cite{melchiorri2000slip}, \cite{su2015force}, \cite{james2020slip} & \cite{bekiroglu2013probabilistic}, \cite{li2018slip}, \cite{calandra2018more}, \cite{hogan2018tactile}\\

\end{tabular}
\label{tab:re2}
\end{table}

\section{Vision-guided tactile perception for crack reconstruction} \label{1.6}
By reviewing the above research on tactile object perception, it is found that many robotic tasks can only be accomplished with vision and tactile sensing together. To illustrate the coordination between visual and tactile perception, we present our work on vision-guided tactile perception for crack detection and reconstruction, with an overview of the framework illustrated in Figure \ref{fig:overview}. 

As for crack detection tasks, skilled inspectors typically first examine the surface to search for areas that have color or shape characteristics similar to cracks, and then use hammers or ultrasonic devices to double-check those areas instead of traversing all regions. 
Inspired by those observations, in this work camera vision is first used for a quick search of the candidate crack regions; a high-resolution optical tactile sensor is then applied against these candidate regions and a refined crack shape is reconstructed from the obtained tactile images.

To evaluate our proposed method, we collected a dataset by investigating 10 mock-up structures that were manufactured using 3D printing. The extensive experiment results demonstrate that compared to only using vision itself, our proposed approach achieves a significant reduction of mean distance error from 0.82 mm to 0.24 mm for crack reconstruction. Furthermore, the proposed method is more than 10 times faster than passive tactile perception in terms of tactile data collection time.

\subsection{Visual Guidance for Touch Sensing} \label{1.3.1}
A deep semantic segmentation network is first utilized to predict pixel-wise masks of cracks. To guide the touch sensing, the contact points with cracks skeletons are then generated. Details of these two steps are given below.


\begin{figure}
\begin{center}

   \includegraphics[width=\linewidth]{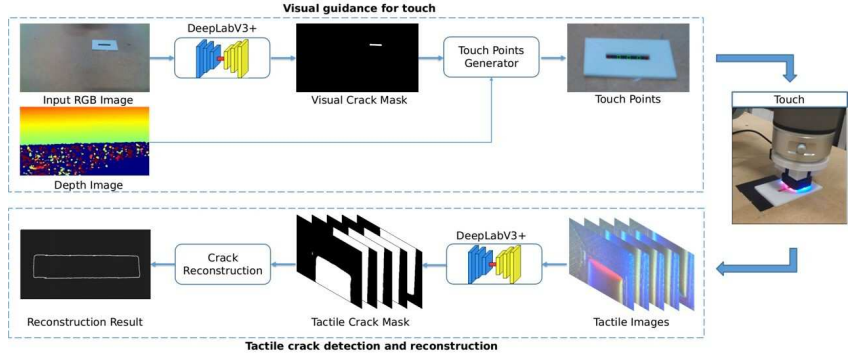}
\end{center}
   \caption{An overview of our vision-guided active tactile crack detection and reconstruction method. \textbf{Top row (from left to right):} The Deeplabv3+ model is used to segment the cracks in the visual image. Given the visual crack mask and the depth image, a set of contact points are generated to guide the collection of tactile images. \textbf{Bottom row (from right to left):} Another deep convolutional network is used to segment the crack in the collected tactile images. Given the detected tactile crack mask, the crack shape are reconstructed based on the geometrical model of the GelSight sensor and the coordinate transforming relation between the tactile  sensor coordinate and the world coordinate. }
\label{fig:overview}
\end{figure}
\textbf{Visual Crack Segmentation.} 
We treat visual crack segmentation as a semantic segmentation problem that classifies pixels of an input image into one of two categories: (a) background (b) cracks. In order to generate these segmentations, we apply the DeepLabv3+ model~\cite{deeplabv3plus2018} that is a state-of-the-art deep learning model for semantic image segmentation. Considering that the number of pixels in the background is much larger than that of the cracks, the network may easily converge to the status that treats all the pixels as background. To address this issue, we use the original images as input instead of resizing it to a smaller size as done in the previous works \cite{pauly2017deeper} and use a weighted cross-entropy loss with crack pixels weighted 10x more than background pixels. In addition, we set the output stride value to 8 since smaller values give finer details in the output mask. To train the model, we started with a pretrained model for semantic segmentation based on the COCO dataset \cite{lin2014microsoft} and use the following hyperparameters: SGD optimizer with a constant learning rate of 1e-6, momentum 0.9 and weight decay 5e-4. 

\textbf{Contact Points Generation.}
Based on the predicted pixel-wise crack mask in the color image, we can extract the crack mask skeleton with a pattern thinning method \cite{zhang1984fast}. To represent the topology of the crack pattern, two types of keypoints (i.e., end points and branch points) and minimal edges are defined as follows: 
\begin{itemize}
    \item  End points: if they have less than two neighbors;  
    \item  Branch points: if they have more than two neighbors; 
    \item  Minimal edge $E_{ij}$: if there is a continuous path between two keypoints $p_i$ and $p_j$ and all points on the path are neither end points nor branch points. 
\end{itemize}

For every minimal edge $E_{ij}$ which consists of a number of ordered points, the keypoint $p_i$ is initially selected as the current contact point $p_{current}$. Then we iteratively choose the next contact point $p_k$ using the following formula: 
\begin{equation}
\begin{array}{l}
\max \limits_{k} D[p_{current},p_k] \\
\text { s.t. } D[p_{current},p_k]<d 
\end{array}
\end{equation}
where $D[p_{current},p_k]$ is the distance between two points in the world coordinate system. The hyper-parameter $d$ is the threshold of the distance between two points that is related to the coverage and speed of the tactile exploration and perception. A smaller $d$ will increase the coverage while reducing the perception speed. In our case, $d$ is empirically set to four fifths of the tactile sensor's view length. As shown in Figure \ref{fig:overview}, the end-point pixels and the generated contact points are tagged with red dots and green dots, respectively. For each contact points $p_i$, the yaw angle of the end-effector is parallel to the vector $<p_i,p_n>$, where $p_n$ is the nearest contact point to $p_i$, so that the end-effector can contact the surface perpendicularly.

\subsection{Guided Tactile Crack Perception}
\textbf{Tactile Crack Detection.}
To address the issue of false positives in the visual crack detection caused by light changes and shadows.
we apply tactile information to refine the vision-based detection results and reconstruct crack shapes in 3D space. 
At first, we control a robot arm with an embedded camera-based Gelsight tactile sensor to collect tactile images autonomously at the generated contact points in Section~\ref{1.3.1}.

\begin{figure}[h]
\begin{center}
   \includegraphics[width=0.6\linewidth]{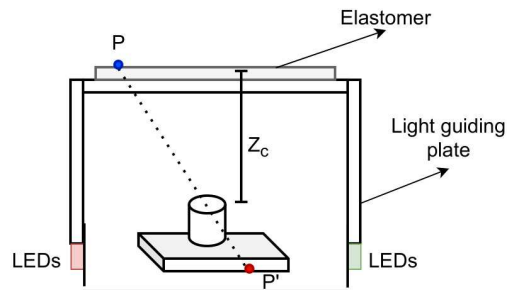}
\end{center}
   \caption{The geometrical model of the GelSight sensor. The webcam at the bottom captures the deformations of the elastomer and the LEDs project light to illuminate the space. }
\label{fig:model}
\end{figure}

The GelSight sensor is a camera-based optical tactile sensor that can capture fine details of the object surface. As shown in Figure~\ref{fig:model}, a webcam under an elastomer captures the deformation of the elastomer as it interacts with the object. The sensor has a flat surface and a view range of $14 mm \times 10.5 mm$ and can capture tactile images at a frequency of 30 Hz \cite{cao2020spatio}.

After collecting the tactile images, another Deeplabv3+ convolutional network \cite{deeplabv3plus2018} with ResNet-101 \cite{he2016deep} is then used to predict the crack masks in tactile images. Since the number of background pixels is similar to that of the crack pixels in tactile images, we use the vanilla cross-entropy loss instead of the weighted cross-entropy loss in Section~\ref{1.3.1}.
Using the predicted masks in tactile images, we can double-check the visual segmentation results. In case there is more than one tactile image with a predicted crack area smaller than a predetermined threshold, all minimal edges will be classified as false positives and deleted. The threshold is empirically set as one-fiftieth of the total number of pixels in the tactile image.

\textbf{Tactile Crack Reconstruction.}
For crack perception, it is crucial to estimate the shape and size of the crack when assessing its potential risk to the building and infrastructure. Due to the limitation of the depth camera's accuracy, current vision-based reconstruction techniques cannot reconstruct small cracks precisely. Therefore, we use the tactile images obtained from the GelSight sensor for the reconstruction of cracks, whose spatial resolution is approximately 20 to 30 microns.

Using the detected boundaries of pixel-wise masks in the tactile images, we first predict the location of cracks on the surface of the GelSight sensor.  To simplify the problem, the webcam is modeled as a pinhole camera, and the surface of the GelSight sensor is treated as a flat plane perpendicular to the camera's $z$ axis. Hence, the transformation between a contact point $P=[X_{c}, Y_{c}, Z_{c}]^{T}$ in the tactile sensor coordinates (the tactile sensor take the centre of the webcam as the origin) to the pixel $P'=[u, v]^{T}$ in the tactile image coordinates can be calculated:

\begin{equation}Z_{c}
\left[\begin{array}{l}
u \\
v \\
1
\end{array}\right]=K
\left[\begin{array}{c}
X_{c} \\
Y_{c} \\
Z_{c} \\
1
\end{array}\right]
\end{equation}
where $Z_c$ is the distance from the optical center to the elastomer surface of the tactile sensor. $K$ is the matrix of the intrinsic parameters of the webcam and can be represented as:
\begin{equation}
K=\left[\begin{array}{llll}
\frac {f}{d_{x}} & 0 & u_{0} & 0 \\
0 & \frac {f}{d_{y}} & v_{0} & 0 \\
0 & 0 & 1 & 0
\end{array}\right]
\end{equation}
where $f$ is the focal length of the webcam, $d_{x}$ and $d_{y}$ denote the pixel size, and $(u_{0}, v_{0})$ is the center point of the tactile image.

After obtaining the position $P$ in the tactile sensor coordinates, we can calculate its position $P_W$ in the world coordinate system:
\begin{equation}
P_{W}=T_{E}^{W} T_{C}^{E} P
\end{equation}
where $T_{C}^{E}$ and $T_{E}^{W}$ are the transformation matrix from the tactile sensor coordinates to the end-effector coordinate system, and from the end-effector coordinate system to the world coordinate system, respectively.

\subsection{Experiment Setup}
\begin{figure}
\begin{center}
   \includegraphics[width=1\linewidth]{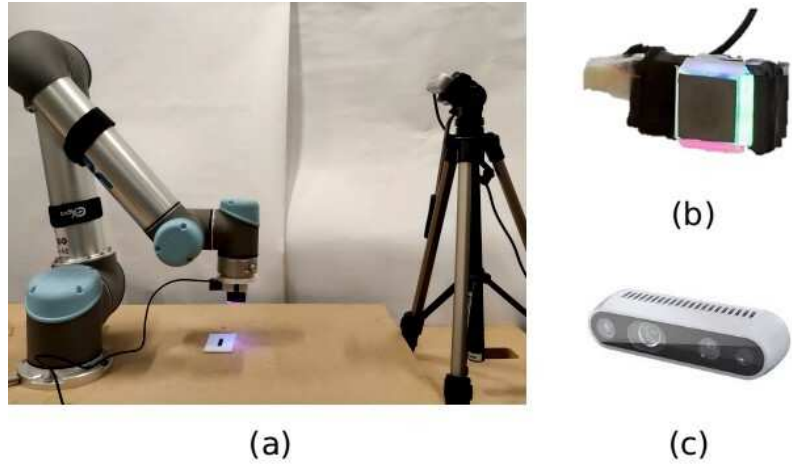}
\end{center}
   \caption{ a) The UR5 robot setup with GelSight sensor and Realsense D435i; b) GelSight sensor; c) Realsense D435i.}
\label{fig:robotsetup}
\end{figure}

In this subsection, we introduce the robot setup used for data collection and experiments with an overview shown in Figure \ref{fig:robotsetup}.

\begin{figure}[h]
\begin{center}
   \includegraphics[width=\linewidth]{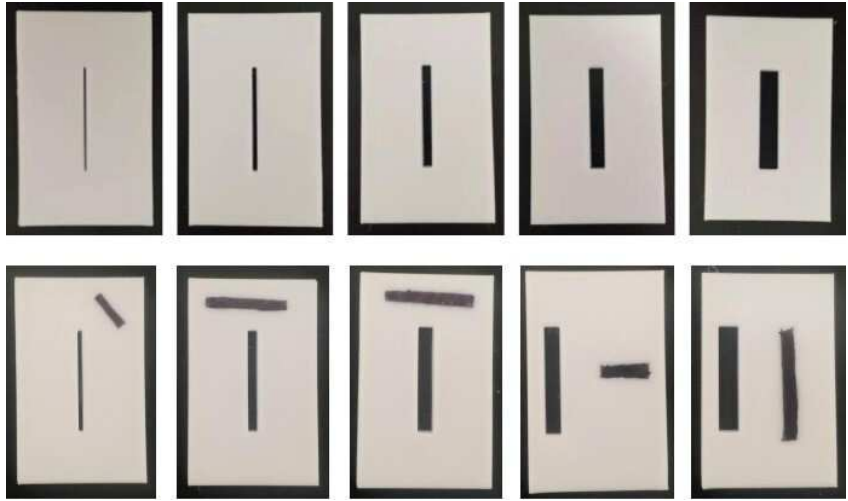}
\end{center}
   \caption{Sample structures used for collecting the visual and tactile dataset. \textbf{Top row}: printed structures with real cracks (holes). \textbf{Bottom row}: printed pad with fake cracks (painted black blocks).}
\label{fig:samples}
\end{figure}

\textbf{3D Printed Structures with Cracks.}
Following part of the data acquisition protocol In~\cite{palermo2020implementing},
a set of 10 structures with cracks of different widths (holes in the structures) are manufactured with PLA plastic. To test the robustness of the proposed method, several structures are painted with fake cracks on the surfaces. The samples are shown in Figure \ref{fig:samples}. 

\textbf{Visual Data Collection.}
For training the visual model, we put those 3D-printed structures on a table and took 10 images of each one containing only real cracks. Then we take 3 images of each object separately for the test. All the images are taken with a RealSense D435i camera.

\begin{figure}
\begin{center}
   \includegraphics[width=\linewidth]{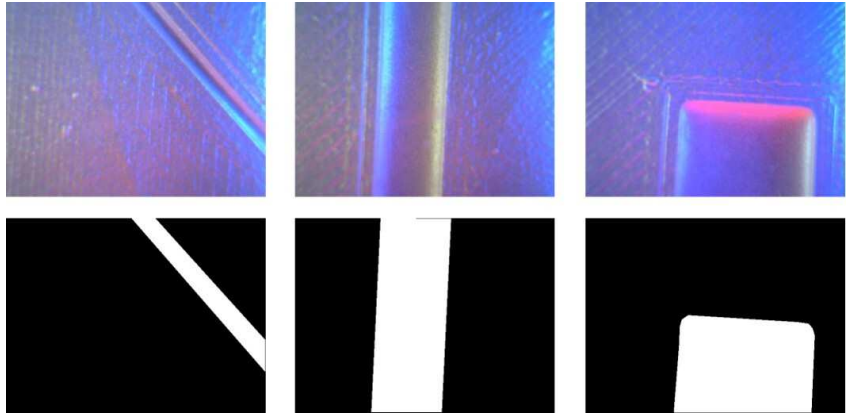}
\end{center}
   \caption{Visualization of the tactile images and their annotations for different cracks. }
\label{fig:tactileimages}
\end{figure}

\textbf{Tactile Data Collection.}
The tactile data collection setup consists of two parts: a 6-DOF Universal Robots UR5 collaborative robot arm and a GelSight sensor mounted on a 3D-printed end effector. To collect the tactile data autonomously and repeatedly, we build a data collection software based on Robot Operating System (ROS). Using the software, a robot arm can be controlled to move across a structure following the pre-determined initial position, step length, and steps in $x$ and $y$ axes. In each position, we rotate the sensor with different angles about the axis perpendicular to the surface in order to generalize the dataset. When the pressure reaches a threshold, the robot arm will stop moving and the GelSight will record a tactile image. In this way, we can obtain high-quality tactile data and avoid stopping the robot arm for unnecessarily protective reasons. In total, 544 valid tactile images were collected and split into training and test sets (370 and 174 for each, respectively), with some samples and their annotations shown in Figure \ref{fig:tactileimages}.  

\subsection{Experiment Results}
\textbf{Detection Results.}  Our proposed methods are evaluated using standard evaluation metrics of pixel accuracy (pixAcc) and Intersection over Union (IoU) . The results of segmentation based solely on visual information and based both on visual and tactile information are summarized in Table \ref{tab:re4}. Due to the presence of fake painting cracks, both pixel accuracy and IoU of visual semantic segmentation have dropped significantly from 0.899 to 0.866 and from 0.504 to 0.376, respectively. In addition, the results show that the crack detection can be effectively improved after tactile information is used to find the fake paintings and refine the visual results. Due to the fact that cracks on the RGB images only take up fewer pixels compared to the background, the pixel accuracy does not change dramatically as IoU.
\begin{table}
	\centering
		\caption{Crack Detection Accuracy.}
		\label{tab:re4}
        \scalebox{1}{
		\begin{tabular}{c| c | c | c }
			\hline
			Modalities & Fake Painting & pixAcc & IoU  \\
			\hline
			\hline
    		vision     &  $\times$ & 0.899 & 0.504\\
    		vision     & \checkmark & 0.866 & 0.376\\
    		vision-tactile  & \checkmark & \textbf{0.909} & \textbf{0.636}\\
    		\hline
		\end{tabular}}
\end{table}

\textbf{Reconstruction Results.} To evaluate the accuracy of our proposed crack reconstruction method, we used the mean, maximum, and Standard Deviation (SD) of the shortest distances between the actual crack shape and the reconstructed crack location. There are four methods used for comparison. The vision method uses point clouds obtained from visual detection and depth information to represent cracks. Aligned vision reduces the impact of depth information accuracy on reconstruction by projecting the point cloud to the table surface. Passive tactile methods collect tactile images by traversing the entire surface of a 3D printed object.

\begin{figure}[h]
\begin{center}
   \includegraphics[width=0.85\linewidth]{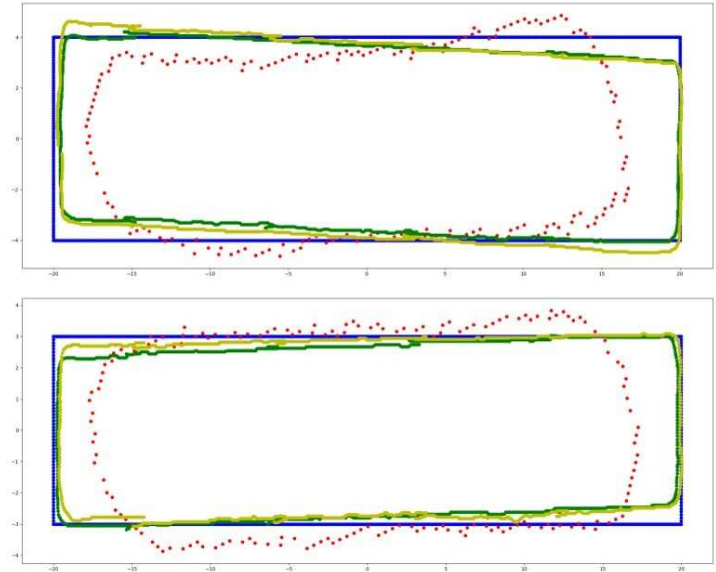}
\end{center}
   \caption{Visual comparison of different reconstruction methods. Blue, red, yellow, green curves represent the ground truth of the crack profile, aligned-vision, passive-tactile and our method for crack reconstruction, respectively.}
\label{fig:reconstruction}
\end{figure}

Table \ref{tab:re5} shows that our approach significantly improves the mean distance error from 0.55 mm to 0.24 mm for crack reconstruction compared to aligned vision method. Two examples of reconstructed crack profiles are shown in Figure \ref{fig:reconstruction}, which shows that reconstructed crack profiles with tactile data are much closer to the ground truth compared to the vision-based method. Furthermore, our proposed vision-guided tactile perception is 10 times faster than passive tactile perception in terms of the time it takes to collect tactile data without affecting the accuracy of crack reconstruction.

\begin{table}
	\centering
		\caption{Reconstruction Accuracy}
		\label{tab:re5}
        \scalebox{1}{
		\begin{tabular}{c| c | c | c | c }
			\hline
			Method & MeanD(mm) & SD(mm) & MaxD(mm) & time(s)  \\
			\hline
			\hline
    		vision & 0.82 & 0.92 & 4.87 & \textbf{1} \\ 
    	    aligned-vision & 0.55 & 0.53 & 3.78 & \textbf{1} \\ 
    		passive-tactile  & \textbf{0.20} & 0.17& 0.99 & 400\\
    		active-tactile[ours]     & 0.24  &\textbf{0.16} & \textbf{0.82} & 35\\
  		\hline
		\end{tabular}}
\end{table}

\section{Conclusion and Discussions}
In this chapter, we first briefly illustrate the irreplaceability of tactile sensing. Then we review the state-of-the-art tactile perception of object properties from four perspectives: material properties, object shape, object pose and grasping stability. A recent trend shows that deep learning, especially CNNs, has been widely applied in robotic tactile perception and optical tactile sensors play a greater role than traditional force tactile sensors thanks to the quick development of both optical sensors and the computer vision field. 
Moreover, we introduce a novel vision-guided tactile perception for crack detection and reconstruction. The cooperation between those two modalities addresses the false positives in visual detection results. The experiment results show that our proposed method can reconstruct crack shapes accurately and efficiently. It has the potential to enable robots to inspect and repair the concrete infrastructure.

As for object property perception using tactile sensing, however, there still exist multiple open issues expected to be investigated: 1) how to achieve real-time perception with optical tactile sensors; 2) how to acquire the costly data with labeling and ground truth; 3) how to close the perception-action loop with tactile sensing. Those open issues and future directions are discussed as follows:

\textbf{Real time tactile perception}.
Due to the development of optical tactile sensor technologies and the domination of deep learning in the computer vision field, CNNs are investigated to achieve better performance at the expense of high computational load. To this end, lighter networks obtained via methods like network pruning can be also considered in those tasks that requires real-time processing in the future.

\textbf{Data acquirement and Labeling}.
It is challenging to obtain ground truth for tactile perception tasks due to the high cost and difficulties in labeling or measuring tactile properties of objects. Most of the recent literature still uses a limited number of human annotators to label data which may bring bias to the ground truths. To this end, unsupervised learning and sim-to-real transfer learning~\cite{gomes2021generation,gomes2019gelsight} will be of benefit in the future research of object perception using tactile sensing. 

\textbf{Close the perception-action loop with tactile sensing}. In the physical interactions with an object, humans  can not only infer object properties but also adjust the hand posture to better feel features that are relevant to the current task. However, most of the studies in the field still focus on research on a dataset collected in advance, whereas the actions taken to collect the data is not considered in the design of the algorithms. To close the perception-action loop, it will be an interesting direction to generate synergies between perception and actions when robots perform complex tasks with tactile sensing~\cite{lu2019surface}. 

 



\Backmatter

 \chapter*{References}
 \markboth{References}{References}
 \bibliographystyle{elsarticle-num} 
 \bibliography{book}


\end{document}